\documentclass{article}

\PassOptionsToPackage{numbers,compress}{natbib}

\usepackage[preprint]{neurips_2026}
\usepackage[utf8]{inputenc}
\usepackage[T1]{fontenc}
\usepackage{hyperref}
\usepackage{url}
\hypersetup{
  colorlinks=true,
  linkcolor=blue,
  citecolor=blue,
  urlcolor=blue
}
\usepackage{booktabs}
\usepackage{amsfonts}
\usepackage{nicefrac}
\usepackage{microtype}
\usepackage[table]{xcolor}
\usepackage{graphicx}
\usepackage{float}
\usepackage{subcaption}
\usepackage{amsmath}
\usepackage{amssymb}
\usepackage{algorithm}
\usepackage{algpseudocode}
\usepackage{mathtools}
\usepackage{amsthm}
\usepackage{pifont}
\usepackage{enumitem}
\usepackage{tabularx}
\usepackage{makecell}
\usepackage{array}
\usepackage{tcolorbox}
\tcbuselibrary{breakable,skins}
\usepackage{fancyvrb}
\usepackage{listings}

\newtcolorbox{promptbox}[1]{
  enhanced,
  breakable,
  colback=yellow!6,
  colframe=yellow!45!black,
  boxrule=0.4pt,
  arc=1mm,
  left=1mm,
  right=1mm,
  top=1mm,
  bottom=1mm,
  before skip=0.35em,
  after skip=0.45em,
  title={\footnotesize\textbf{#1}},
  fonttitle=\footnotesize,
  coltitle=black
}

\DefineVerbatimEnvironment{PromptVerbatim}{Verbatim}{
  fontsize=\scriptsize
}

\usepackage[capitalize,noabbrev]{cleveref}

\theoremstyle{plain}

\theoremstyle{definition}

\theoremstyle{remark}

\title{\emph{EvoMAS}: Learning Execution-Time Workflows for Multi-Agent Systems}

\author{
\textbf{Chengdong Xu}{\footnotesize $^\bigstar$} \quad
\textbf{Kaiqiang Ke}{\footnotesize $^\bigstar$} \quad
\textbf{Ziheng Liu}{\footnotesize $^\diamondsuit$} \quad
\textbf{Jiaqi Wei}{\footnotesize $^\spadesuit$} \\
\textbf{Zibo Shao}{\footnotesize $^\heartsuit$} \quad
\textbf{Weile Guo}{\footnotesize $^\bigstar$} \quad
\textbf{Chao Yu}{\footnotesize $^\bigstar$\textsuperscript{\ding{41}}} \\[0.3em]
{\footnotesize $^\bigstar$}Sun Yat-Sen University \quad
{\footnotesize $^\spadesuit$}Zhejiang University \\
{\footnotesize $^\diamondsuit$}ShanghaiTech University \\
{\footnotesize $^\heartsuit$}Institute of Automation, Chinese Academy of Sciences \\[0.3em]
\texttt{xuchd6@mail2.sysu.edu.cn} \quad
}

\begin{document}

\maketitle
\renewcommand{\thefootnote}{}
\footnotetext{\textrm{\ding{41}} Corresponding author.}
\renewcommand{\thefootnote}{\arabic{footnote}}

% \begin{abstract}
% Large language model (LLM)-based multi-agent systems have demonstrated strong performance on complex tasks by leveraging agent specialization, tool use, and structured collaboration. However, most automated multi-agent design methods still follow a one-shot paradigm, where coordination structures are determined before execution and kept fixed throughout the task. This is ill-suited for long-horizon problems that unfold over multiple stages with evolving subgoals, intermediate results, and information needs.
% We propose \textbf{EvoMAS}, a framework for \emph{execution-time multi-agent workflow construction} that formulates agent coordination as sequential decision-making along a single task trajectory. At each stage, EvoMAS constructs a structured task state (via a Planner, an Evaluator, and an Updater) and uses a learned Workflow Adapter to build a stage-specific workflow from a fixed pool of candidate agents.
% The workflow construction policy is optimized with policy gradients using a sparse, verifiable terminal reward defined by task-level correctness, with the final utility uniformly assigned to all stages. Experiments on complex multi-step tasks show that execution-time, state-conditioned workflow construction yields more robust and effective multi-agent behavior than static initialization-time designs.
% \end{abstract}

\begin{abstract}
    Large language model (LLM)-based multi-agent systems have shown strong
    potential on complex tasks through agent specialization, tool use, and
    collaborative reasoning. However, most automated multi-agent system design
    methods still follow a one-shot paradigm: a workflow is optimized or selected
    before execution and then reused unchanged throughout the task. This static
    coordination strategy is ill-suited for long-horizon tasks whose subgoals,
    intermediate evidence, and information needs evolve over multiple execution
    stages.
    We propose \textbf{EvoMAS}, a framework for \emph{execution-time multi-agent
    workflow construction}. EvoMAS formulates workflow construction as a
    meta-level sequential decision problem along a single task trajectory. At each
    stage, it constructs an explicit task state through a Planner--Evaluator--Updater
    pipeline and uses a learned Workflow Adapter to instantiate a stage-specific
    layered workflow from a fixed pool of candidate agents. The adapter is trained
    with policy gradients using sparse, verifiable terminal task success as the
    main supervision signal, while evaluator-based process reward is analyzed
    separately under very-hard sparse-reward settings.
    Experiments on GAIA, HLE, and DeepResearcher show that EvoMAS outperforms
    single-agent baselines and recent automated multi-agent workflow design
    methods. Our analyses further show that explicit task-state construction and
    learned workflow adaptation provide complementary benefits. Additional results
    indicate that process reward is most useful when terminal success is extremely
    sparse, and qualitative case studies illustrate that EvoMAS adapts agent
    coordination as the task state evolves.
    \end{abstract}

\section{Introduction}

Large language model (LLM)-based agents~\citep{autogpt,babyagi,metagpt} extend pure language
generation with capabilities such as tool use~\citep{react,toolformer,hugginggpt,gorilla,toollm,apibank,restgpt},
planning~\citep{planandsolve,treeofthoughts}, and contextual reasoning~\citep{zeroshotcot}, enabling strong empirical
performance across tasks including question answering, data analysis, code generation, and web interaction~\citep{webgpt,gpt4,codex,webarena,mind2web}.
Building on single-agent systems, multi-agent frameworks further improve
performance through specialization, parallel reasoning, and mutual verification
among agents~\citep{autogen,metagpt,NeurIPS2023camel,llmdebate,agentverse}, offering a promising approach for solving
complex tasks that exceed the capacity of monolithic prompting or single-pass
inference.

Early LLM-based multi-agent systems, such as
AutoGen~\citep{autogen}, MetaGPT~\citep{metagpt}, and CAMEL~\citep{NeurIPS2023camel}, demonstrated the benefits of
agent collaboration but relied on manually specified prompts, roles, and communication pipelines~\citep{wang2023agentsurvey}.
Subsequent work has sought to automate these design choices, including optimizing
prompts, agent profiles, coordination structures, and execution workflows~\citep{gptswarm,aflow,agentprune,evomac,evoagent,gdesigner,maas,evolvingorchestration,verimaas}.
Despite these advances, most existing approaches adopt a \emph{one-shot
initialization} paradigm: the multi-agent workflow is designed prior to execution
and then reused without modification throughout the task~\citep{gptswarm,aflow,agentprune,gdesigner,maas}.

This paradigm becomes increasingly inadequate for \emph{complex, multi-step, and
long-horizon tasks}, which typically require multiple execution stages with
shifting subgoals, intermediate artifacts, and changing information requirements~\citep{gaia,webarena,hle,deepresearcher,browsecomp}.
In such settings, relying on a single workflow with a fixed coordination strategy
is inherently suboptimal~\citep{gaia,wang2023agentsurvey}.
To support execution on such complex tasks,
a multi-agent system must make \emph{stage-wise} coordination decisions during execution, rather than committing to a single workflow upfront.
Achieving this paradigm requires addressing two fundamental challenges.

\textbf{Challenge 1 (Constructing an Explicit Execution-Time Task State).}
Long-horizon tasks evolve over multiple execution stages, involving changing subgoals,
accumulating intermediate results, and potential failures that require revision
or refinement~\citep{gaia,webarena,mind2web}.
To make principled coordination decisions at execution time, a system must maintain
an explicit task state that summarizes execution progress and solution quality~\citep{wang2023agentsurvey}.
Without such a state representation, workflow decisions lack a concrete basis for
adapting to the current execution context.

\begin{figure*}[t]
    \centering
    \includegraphics[width=1\linewidth]{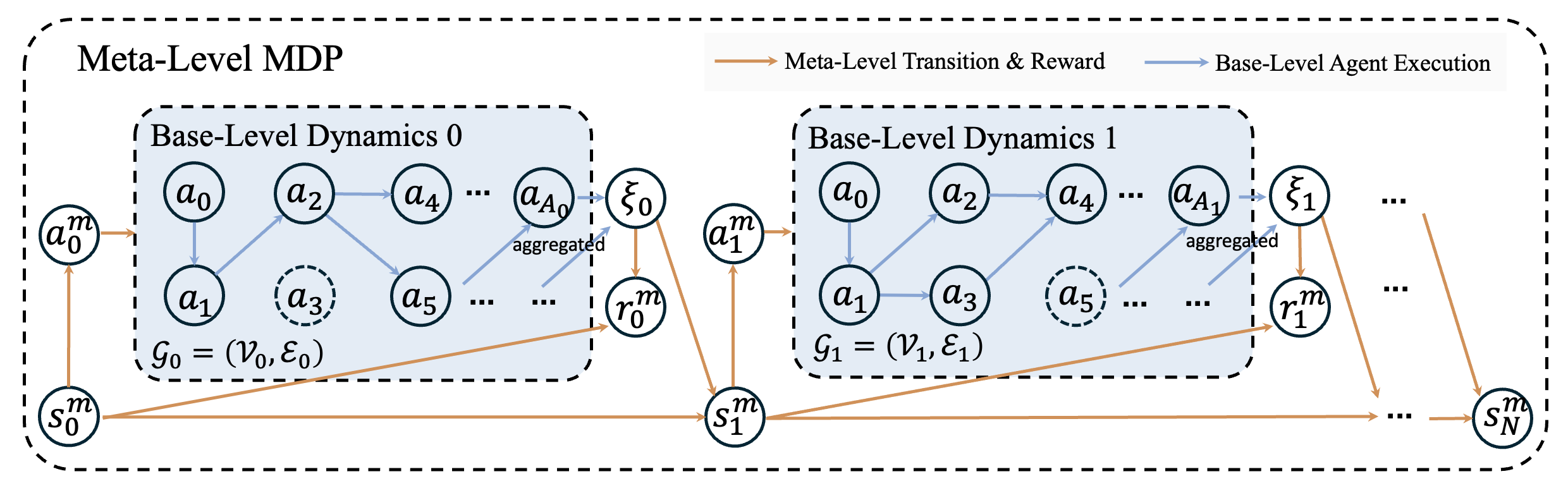}
    \caption{\textbf{Meta-level MDP formulation of execution-time multi-agent workflow construction.}
A workflow $\mathcal{G}_t$ is selected based on the meta-state $s_t^{m}$, inducing
base-level multi-agent execution and an aggregated outcome $\xi_t$.
Orange arrows denote meta-level transitions, and blue arrows denote base-level agent execution.}
    % \vspace{-0.5em}
    \label{fig:meta_mdp}
\end{figure*}

\textbf{Challenge 2 (Adapting Workflows to Evolving Task States).}
Even with an explicit task state, effective coordination requires the ability to
adapt the multi-agent workflow as execution progresses.
Different execution stages impose heterogeneous coordination requirements:
workflows suited for early exploration may be inefficient or counterproductive
during refinement and verification~\citep{gaia,webarena}.
Thus, a system must dynamically construct stage-specific workflows conditioned on
the current task state, rather than relying on a single static design~\citep{gptswarm,aflow,agentprune,gdesigner,maas}.

To address these challenges, we propose \textbf{EvoMAS}, a framework for
\emph{execution-time multi-agent workflow construction} that treats workflow
selection as a sequential decision-making problem along a single task trajectory.
Unlike methods that optimize or instantiate a single workflow before execution,
EvoMAS makes workflow decisions during execution, conditioned on the evolving
task state.
At each execution stage, EvoMAS constructs an explicit task state that summarizes
execution progress and quality, and dynamically selects a stage-specific
multi-agent workflow conditioned on this state.
The workflow construction policy is learned via reinforcement learning, enabling
execution-aware adaptation of agent coordination structures rather than static,
initialization-time designs.

Our contributions are summarized as follows:
% \vspace{-0.5em}
\begin{itemize}[leftmargin=*,itemsep=-0.2em]
\item[\ding{182}] \textbf{\textit{Paradigm Reformulation}:}
We reformulate automated multi-agent workflow construction as a sequential
decision-making problem during task execution, moving beyond one-shot,
initialization-time designs.
\item[\ding{183}] \textbf{\textit{Execution-Time Framework}:}
We propose EvoMAS, which combines explicit task state construction with a learned,
state-conditioned workflow adapter to dynamically reorganize multi-agent workflows
during execution.
\item[\ding{184}] \textbf{\textit{Empirical Validation}:}
Through experiments on complex, multi-step, and long-horizon tasks,
we compare EvoMAS with single-agent baselines and recent automated
multi-agent workflow design methods, and provide ablations on task state,
workflow learning, reward signals, and backbone capability.
\end{itemize}
\vspace{-0.5em}

\vspace{-0.3em}
\section{Formalization}
\label{sec:formal}
\vspace{-0.3em}

We formulate execution-time workflow construction as a meta-level Markov
decision process, as illustrated in Figure~\ref{fig:meta_mdp}. Let $\tau$
denote a task instance and $\mathcal{P}=\{a_1,\ldots,a_N\}$ denote a fixed
pool of candidate agents. At each meta-step $t$, the system maintains a
meta-state $s_t^m$, which summarizes the current task progress, intermediate
artifacts, and execution feedback. The system then selects a meta-action
$a_t^m$, which corresponds to a concrete multi-agent workflow
\begin{equation}
a_t^m \equiv \mathcal{G}_t=(\mathcal{V}_t,\mathcal{E}_t)\in\mathcal{W},
\qquad \mathcal{V}_t\subseteq \mathcal{P},
\end{equation}
where $\mathcal{V}_t$ is the selected agent set, $\mathcal{E}_t$ specifies the
information flow, and $\mathcal{W}$ denotes the feasible space of layered DAG
workflows over the agent pool. In our implementation, $\mathcal{V}_t$ is
organized into $L_t$ workflow layers (with $L_t\le L$ for a fixed maximum
depth),
$\mathcal{V}_t=\bigcup_{\ell=1}^{L_t}\mathcal{V}_t^{(\ell)}$, and
$\mathcal{E}_t$ connects agents across consecutive layers.

Formally, the meta-level process is defined as
\begin{equation}
\mathcal{M}=(\mathcal{S}^m,\mathcal{A}^m,\mathcal{T}^m,\mathcal{R}^m),
\end{equation}
where $\mathcal{S}^m$ is the meta-state space and the meta-action space is the
workflow space, i.e., $\mathcal{A}^m\equiv\mathcal{W}$. A workflow policy
selects
\begin{equation}
a_t^m \sim \pi_{\mathrm{wf}}(\cdot\mid \tau,s_t^m).
\end{equation}
Executing the selected workflow $\mathcal{G}_t$ induces base-level multi-agent
dynamics and produces an aggregated outcome $\xi_t$, including agent messages,
tool observations, and evaluator feedback. The meta-state transition is then
\begin{equation}
s_{t+1}^m \sim
\mathcal{T}^m(\cdot\mid \tau,s_t^m,a_t^m,\xi_t),
\end{equation}
and the corresponding meta-level reward is
\begin{equation}
r_t^m=\mathcal{R}^m(\tau,s_t^m,a_t^m,\xi_t).
\end{equation}
In the main experiments, we use sparse terminal task success as the trajectory
utility; evaluator-based process reward is analyzed separately in
Section~\ref{sec:additional_analysis}.

The objective is to learn a state-conditioned workflow policy that maximizes
expected cumulative meta-level reward:
\begin{equation}
\max_{\theta}
\mathbb{E}_{\pi_{\mathrm{wf},\theta}}
\left[
\sum_{t=1}^{T} r_t^m
\right].
\end{equation}

\begin{figure*}[t]
    \centering
    \includegraphics[width=1\linewidth]{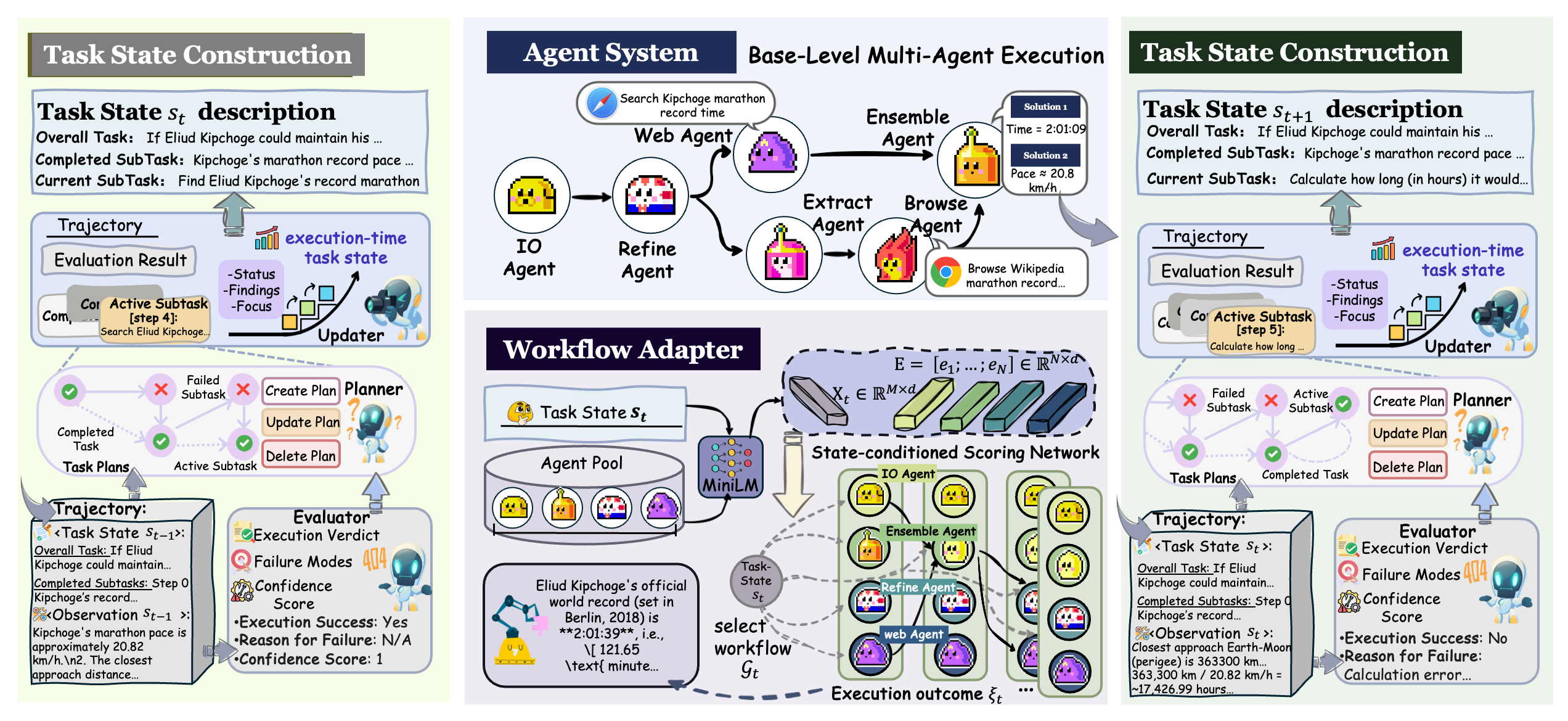}
    \vspace{-1em}
    \caption{\textbf{Overview of EvoMAS.} The system alternates between execution-time task state construction and state-conditioned workflow adaptation, enabling multi-agent coordination to evolve along a single task trajectory.}
    \label{fig:framework_overview}
    % \vspace{-0.5em}
\end{figure*}

\section{EvoMAS}
% \vspace{-0.4em}
\label{sec:method}

EvoMAS instantiates the Meta-MDP in Section~\ref{sec:formal} with three
components: task-state construction, state-conditioned workflow construction,
and workflow-policy optimization. Given a task $\tau$, EvoMAS proceeds over
meta-steps: it constructs a structured task state $\tilde{s}_t$ and then
selects a stage-specific workflow $\mathcal{G}_t$ from a fixed candidate agent
pool. The workflow policy is learned with reinforcement learning so that
coordination adapts as the task state evolves (\Cref{fig:framework_overview}).

% \vspace{-0.4em}
\subsection{Task State Construction}
\label{subsec:state_construction}
% \vspace{-0.4em}

EvoMAS represents the meta-state $s_t^m$ as a structured textual task state
$\tilde{s}_t$. This state summarizes the current execution progress and serves
two purposes: it conditions workflow selection and provides execution-time
working context. At each stage, $\tilde{s}_t$ is constructed by a
Planner--Evaluator--Updater pipeline.

The \textbf{Planner} maintains the current subtask plan and identifies the
active subtask. If the previous execution succeeds, the Planner advances to
the next pending subtask; if evidence is missing or execution fails, it may
keep the current subtask and attach the failure context for refinement. This
supports both task decomposition and iterative correction.

The \textbf{Evaluator} assesses the previous execution outcome $\xi_{t-1}$
with respect to the active subtask. It produces stage-level assessment signals
such as a success/refinement verdict, confidence, and natural-language
feedback, which are part of the aggregated execution record $\xi_{t-1}$. In
the main training setting, these signals are used for state
construction rather than as direct optimization objectives; evaluator-based
process reward is studied separately in Section~\ref{sec:additional_analysis}.

The \textbf{Updater} implements the meta-state transition by integrating the
previous task state, the executed workflow, and the aggregated execution
record:
\begin{equation}
\tilde{s}_t =
\mu(\tau,\tilde{s}_{t-1},\mathcal{G}_{t-1},\xi_{t-1}).
\end{equation}
The resulting state contains the overall task objective, the active subtask,
completed subtasks with salient intermediate artifacts, and feedback useful
for subsequent decisions. EvoMAS uses a compact serialization of $\tilde{s}_t$
to condition the Workflow Adapter, while the full structured state is provided
to agents as execution-time context. Thus, $\mu$ implements the meta-level
transition $\mathcal{T}^m$ in Section~\ref{sec:formal}.

% \vspace{-0.4em}
\subsection{Execution-Time Workflow Construction}
\label{subsec:design}
% \vspace{-0.4em}

Given the task state $\tilde{s}_t$, EvoMAS constructs a stage-specific
multi-agent workflow rather than reusing a fixed workflow throughout the task.
The Workflow Adapter operates over a fixed candidate agent pool
$\mathcal{P}=\{a_1,\ldots,a_N\}$ and instantiates a layered workflow
\begin{equation}
\mathcal{G}_t=(\mathcal{V}_t,\mathcal{E}_t),
\qquad
\mathcal{V}_t=\bigcup_{\ell=1}^{L_t}\mathcal{V}_t^{(\ell)},
\end{equation}
where $\mathcal{V}_t^{(\ell)}\subseteq\mathcal{P}$ denotes the agents selected
at layer $\ell$. Consecutive layers are connected by fully connected
bipartite edges:
\begin{equation}
\mathcal{E}_t =
\{(a_i,a_j)\mid a_i\in\mathcal{V}_t^{(\ell)},
a_j\in\mathcal{V}_t^{(\ell+1)},\ell=1,\ldots,L_t-1\}.
\end{equation}
Each downstream agent receives an aggregated message from all agents in the
previous layer. The sampling procedure is applied separately at each layer,
so the same agent may appear in multiple layers and across different
meta-steps.

To condition workflow construction on the current execution context, EvoMAS
encodes a compact serialization of $\tilde{s}_t$ into token representations
$\mathbf{X}_t\in\mathbb{R}^{M\times d}$. Each candidate agent $a_i$ is
represented by an embedding $\mathbf{e}_i\in\mathbb{R}^d$ derived from its role,
capabilities, and tool interface. At layer $\ell$, the Workflow Adapter uses
cross-attention between the task-state representation and agent embeddings to
produce a layer-specific query vector $\mathbf{h}_t^{(\ell)}$, scoring agents by
\begin{equation}
\alpha_{t,i}^{(\ell)}
=
\cos(\mathbf{h}_t^{(\ell)},\mathbf{e}_i).
\end{equation}
We convert scores into a categorical distribution and soft-sample with a
cumulative probability threshold:
\begin{equation}
\mathbf{p}_{t}^{(\ell)}
=
\mathrm{softmax}\!\left(\boldsymbol{\alpha}_{t}^{(\ell)}/\lambda\right),
\qquad
p_{t,i}^{(\ell)}
=
\frac{\exp(\alpha_{t,i}^{(\ell)}/\lambda)}{\sum_{j=1}^{N}\exp(\alpha_{t,j}^{(\ell)}/\lambda)},
\end{equation}
where $\lambda$ is a temperature parameter. We then sample agents \emph{without
replacement} from $p_{t}^{(\ell)}$ until the cumulative probability mass of
the selected set reaches a threshold $\rho\in(0,1]$. Denoting the resulting
random index set by $S_t^{(\ell)}$, we define
\begin{equation}
\mathcal{V}_t^{(\ell)}=\{a_i\mid i\in S_t^{(\ell)}\}.
\end{equation}
Repeating this procedure for $\ell=1,\ldots,L_t$ yields the workflow
$\mathcal{G}_t$ executed at meta-step $t$.

% \vspace{-0.4em}
\subsection{Workflow Policy Optimization}
\label{subsec:optimization}
% \vspace{-0.4em}

EvoMAS optimizes the Workflow Adapter parameters $\Theta$ using sparse
verifiable terminal reward. For a task $\tau$, execution produces a final
answer $\hat{a}(\tau)$, and the task utility $U(\tau)$ is computed by the
benchmark evaluation protocol. We use terminal reward in the main experiments
because it is verifiable and avoids directly optimizing against evaluator
judgments; evaluator-based process reward is studied separately in
Section~\ref{sec:additional_analysis}.

At each meta-step, the Workflow Adapter samples
\begin{equation}
\mathcal{G}_t\sim\pi_{\mathrm{wf}}(\cdot\mid\tilde{s}_t;\Theta).
\end{equation}
The workflow log-probability is defined as the log-probability of the sampled
agent selections under the workflow policy. Let $\mathbf{i}_t^{(\ell)}$ denote
the (ordered) index sequence produced by the layer-$\ell$ sampling procedure.
We define
\begin{equation}
\log P_\Theta(\mathcal{G}_t\mid\tilde{s}_t)
=
\sum_{\ell=1}^{L_t}\log P_\Theta\!\left(\mathbf{i}_t^{(\ell)}\mid\tilde{s}_t\right),
\end{equation}
where $P_\Theta(\mathbf{i}_t^{(\ell)}\mid\tilde{s}_t)$ is induced by our
cumulative-mass soft-sampling rule (Appendix~\ref{app:workflow_adapter}).
Although the executed workflow depends only on the selected set
$\mathcal{V}_t^{(\ell)}$, we record the sampled order $\mathbf{i}_t^{(\ell)}$
and use its probability for policy-gradient training.
The objective is
\begin{equation}
\mathcal{J}(\Theta)
=
\mathbb{E}_{\tau\sim\mathcal{Q}}
\mathbb{E}_{\{\mathcal{G}_t\}\sim\pi_{\mathrm{wf}}}
[U(\tau)].
\end{equation}
We optimize it with REINFORCE:
\begin{equation}
\nabla_\Theta\mathcal{J}(\Theta)
\approx
\mathbb{E}
\left[
\sum_{t=1}^{T}
U(\tau)\nabla_\Theta
\log P_\Theta(\mathcal{G}_t\mid\tilde{s}_t)
\right].
\end{equation}
This trains the adapter to prefer workflows that improve terminal task
success under sparse supervision.

% \vspace{-0.4em}
\section{Experiments}
\label{sec:experiments}
% \vspace{-0.4em}

\subsection{Experimental Setup}
% \vspace{-0.4em}
% \paragraph{Benchmarks and Splits.}
% We evaluate EvoMAS on GAIA, HLE, and DeepResearcher, and use GAIA Level 3
% and BrowserComp for additional analysis under very-hard sparse-reward
% settings. GAIA emphasizes multi-step tool use and information gathering, HLE
% provides challenging knowledge-intensive reasoning tasks, and DeepResearcher
% evaluates longer research-style problem solving. BrowserComp is used as a
% diagnostic very-hard setting because its tasks often provide extremely sparse
% terminal success signals.

% For GAIA Level~1 and Level~2, we use all released tasks at the corresponding
% difficulty and randomly partition them into training and validation subsets
% with an 8:2 ratio.
% For HLE and DeepResearcher, we sample 50 instances from the original datasets
% and split them into training and validation sets with the same 8:2 ratio.
% BrowserComp is used only as a small very-hard diagnostic set (20 instances
% with a 16/4 train--validation split).
% Unless otherwise specified, reported results are evaluated on the held-out
% evaluation split used in each experiment. For small diagnostic sets, we report
% exact counts when appropriate to avoid over-interpreting percentage differences.
\paragraph{Benchmarks.}
We evaluate EvoMAS on GAIA, HLE, and DeepResearcher, covering multi-step tool
use, knowledge-intensive reasoning, and research-style problem solving. We
also use GAIA Level~3 and BrowserComp as very-hard diagnostic settings for
analyzing sparse-reward regimes. Unless otherwise specified, results are
reported on held-out evaluation splits; detailed sampling and split
information is provided in Appendix~\ref{app:experimental_details}.

% \paragraph{Agent Pools and Workflow Configuration.}
% EvoMAS constructs 3-layer workflows from a fixed candidate agent pool. We
% consider two pool sizes: EvoMAS-4 uses I/O, Early-exit, Web-search, and
% Ensemble agents, while EvoMAS-7 further includes Multi-generate, Self-refine,
% and Web-browser agents. The Early-exit agent terminates the workflow when the
% current state is sufficient for final answering. Comparing EvoMAS-4 and
% EvoMAS-7 isolates whether execution-time workflow adaptation can benefit from
% a richer set of candidate coordination choices.
\paragraph{Agent Pools and Workflow Configuration.}
EvoMAS constructs 3-layer workflows from a fixed candidate agent pool. We use
two pool sizes: EvoMAS-4 contains I/O, Early-exit, Web-search, and Ensemble
agents, while EvoMAS-7 further adds Multi-generate, Self-refine, and
Web-browser agents. This comparison isolates whether execution-time workflow
adaptation benefits from richer candidate coordination choices.

\paragraph{Baselines.}
We compare against two baseline groups. GPT-4o-mini and GPT-4o serve as strong
single-agent references. GPTSwarm~\cite{gptswarm}, AFlow~\cite{aflow},
G-Designer~\cite{gdesigner}, and MaAS~\cite{maas} represent recent automated
MAS design methods, making them the most direct comparisons for workflow-level
coordination. We focus on these baselines to test whether
execution-time, state-conditioned workflow construction improves over static or
initialization-time automated MAS designs, while avoiding confounds from
hand-crafted, benchmark-specific agent pipelines.
% \paragraph{Training and Evaluation.}
% The main results are trained with sparse terminal reward only, without process
% reward. Task success rate is used as the primary metric. All methods are
% evaluated under the same maximum number of execution stages and tool-call
% budget. For automated MAS baselines, we use the same benchmark splits and
% execution budget whenever applicable. When a baseline does not natively
% support a benchmark-specific tool interface, we instantiate it with the
% closest matching agent/tool configuration and keep the evaluation protocol
% unchanged. Detailed prompts, benchmark splits, model identifiers, and
% cost-statistics templates are provided in the appendix.
\paragraph{Training and Evaluation.}
The main EvoMAS results are trained with sparse terminal reward only, without
process reward. Task success rate is used as the primary metric. All methods
are evaluated under the same maximum number of execution stages and tool-call
budget whenever applicable. Detailed prompts, model identifiers, benchmark
splits, and cost-statistics templates are provided in the appendix.
\subsection{Main Results}
% \vspace{-0.4em}

Table~\ref{tab:main_results} reports the main results. EvoMAS-7 achieves the
best average success rate of 65.0, outperforming the strongest automated MAS
baseline, G-Designer/MaAS, by 30.2 points and the strongest single-agent
baseline, GPT-4o, by 35.2 points. The gains are broad rather than
benchmark-specific: EvoMAS-7 obtains the best result on all four evaluated
benchmarks, including GAIA Level 1, GAIA Level 2, HLE, and DeepResearcher.

The comparison between EvoMAS-4 and EvoMAS-7 further shows the effect of
candidate-pool richness under the same execution-time workflow adaptation
framework. Expanding the pool from 4 to 7 agents improves the average score
from 45.0 to 65.0, with especially large gains on GAIA Level 2 (+28.7), HLE
(+20.0), and DeepResearcher (+20.0). This suggests that harder and more
diverse tasks benefit more from richer execution-time coordination choices.

Notably, the automated MAS baselines do not consistently outperform strong
single-agent baselines across all benchmarks, suggesting that optimizing or
designing a static workflow before execution is insufficient for heterogeneous
long-horizon tasks. EvoMAS differs by adapting the workflow at execution time
based on the evolving task state, allowing the system to shift between
information acquisition, refinement, aggregation, and early termination as
needed.

\begin{table}[t]
    \centering
    \caption{\textbf{Main results on multi-step agent benchmarks.} EvoMAS uses 3 workflow layers and is trained with sparse terminal reward only. EvoMAS-4 and EvoMAS-7 denote the 4-agent and 7-agent candidate pools, respectively.}
    \label{tab:main_results}
    \small
    % tabularx (not tabular*+extracolsep) so \rowcolor spans full width without white gaps between columns
    \definecolor{tblmas}{HTML}{F4FAFC}
    \definecolor{tblsub}{HTML}{F0F6FA}
    \setlength{\tabcolsep}{5pt}
    \renewcommand{\arraystretch}{1.12}
    \begin{tabularx}{\textwidth}{>{\raggedright\arraybackslash}Xccccc}
    \toprule
    \textbf{Method} & \textbf{GAIA Level1} & \textbf{GAIA Level2} & \textbf{HLE} & \textbf{DeepResearcher} & \textbf{Average} \\
    \midrule
    \rowcolor{tblsub}
    \multicolumn{6}{@{}l}{\textit{Single-agent baselines}} \\
    GPT-4o-mini & 22.2 & 7.1 & 10.0 & 50.0 & 22.3 \\
    GPT-4o      & 22.2 & 7.1 & 30.0 & 60.0 & 29.8 \\
    \midrule
    \rowcolor{tblsub}
    \multicolumn{6}{@{}l}{\textit{Automated MAS baselines}} \\
    GPTSwarm   & 33.3 & 0.0 & 20.0 & 20.0 & 18.3 \\
    AFlow      & 11.1 & 7.1 & 20.0 & 50.0 & 22.1 \\
    G-Designer & 22.2 & 7.1 & 50.0 & 60.0 & 34.8 \\
    MaAS       & 22.2 & 7.1 & 50.0 & 60.0 & 34.8 \\
    \midrule
    \multicolumn{6}{@{}l}{\textit{Ours}} \\
    \rowcolor{tblmas}
    EvoMAS-4 & 55.6 & 14.2 & 40.0 & 70.0 & 45.0 \\
    \rowcolor{tblmas}
    \textbf{EvoMAS-7} & \textbf{66.7} & \textbf{42.9} & \textbf{60.0} & \textbf{90.0} & \textbf{65.0} \\
    \bottomrule
    \end{tabularx}
    \vspace{1em}
\end{table}

% \begin{table}[t]
% \centering
% \caption{Main results on multi-step agent benchmarks. EvoMAS uses 3 workflow layers and is trained with sparse terminal reward only. EvoMAS-4 and EvoMAS-7 denote the 4-agent and 7-agent candidate pools, respectively.}
% \label{tab:main_results}
% \small
% \setlength{\tabcolsep}{7pt}
% \renewcommand{\arraystretch}{1.05}
% \begin{tabular}{lccccc}
% \toprule
% \textbf{Method} & \textbf{GAIA L1} & \textbf{GAIA L2} & \textbf{HLE} & \textbf{DeepRes.} & \textbf{Avg.} \\
% \midrule
% \multicolumn{6}{l}{\textit{Single-agent baselines}} \\
% GPT-4o-mini & 22.2 & 7.1 & 10.0 & 50.0 & 22.3 \\
% GPT-4o      & 22.2 & 7.1 & 30.0 & 60.0 & 29.8 \\
% \midrule
% \multicolumn{6}{l}{\textit{Automated MAS baselines}} \\
% GPTSwarm   & 33.3 & 0.0 & 20.0 & 20.0 & 18.3 \\
% AFlow      & 11.1 & 7.1 & 20.0 & 50.0 & 22.1 \\
% G-Designer & 22.2 & 7.1 & 50.0 & 60.0 & 34.8 \\
% MaAS       & 22.2 & 7.1 & 50.0 & 60.0 & 34.8 \\
% \midrule
% \multicolumn{6}{l}{\textit{Ours}} \\
% EvoMAS-4 & 55.6 & 14.2 & 40.0 & 70.0 & 45.0 \\
% EvoMAS-7 & \textbf{66.7} & \textbf{42.9} & \textbf{60.0} & \textbf{90.0} & \textbf{65.0} \\
% \bottomrule
% \end{tabular}
% \vspace{-0.4em}
% \end{table}

\subsection{Task-State Construction and Workflow Learning}
\label{subsec:ablation}
% \vspace{-0.4em}

\paragraph{Does Execution-Time Task State Matter?}
We first ask whether explicitly maintaining an \emph{execution-time task state}
helps long-horizon performance when coordination is otherwise held fixed.
Concretely, we keep the multi-agent workflow identical across conditions and
toggle only the Planner--Evaluator--Updater pipeline that serializes plans,
outcomes, and evaluator feedback into the next structured state.
On GAIA, we use a fixed
\texttt{I/O} $\rightarrow$ \texttt{Web-search} $\rightarrow$ \texttt{Ensemble}
workflow instantiated with GPT-4.1 agents.
On HLE, we use a fixed
\texttt{Multi-generate-CoT} $\rightarrow$ \texttt{Ensemble} reasoning chain and
repeat the comparison under GPT-4o and Gemini-2.5-Pro to test backbone
robustness.

\begin{figure}[t]
\centering
\begin{subfigure}[t]{0.58\linewidth}
    \centering
    \includegraphics[width=\linewidth]{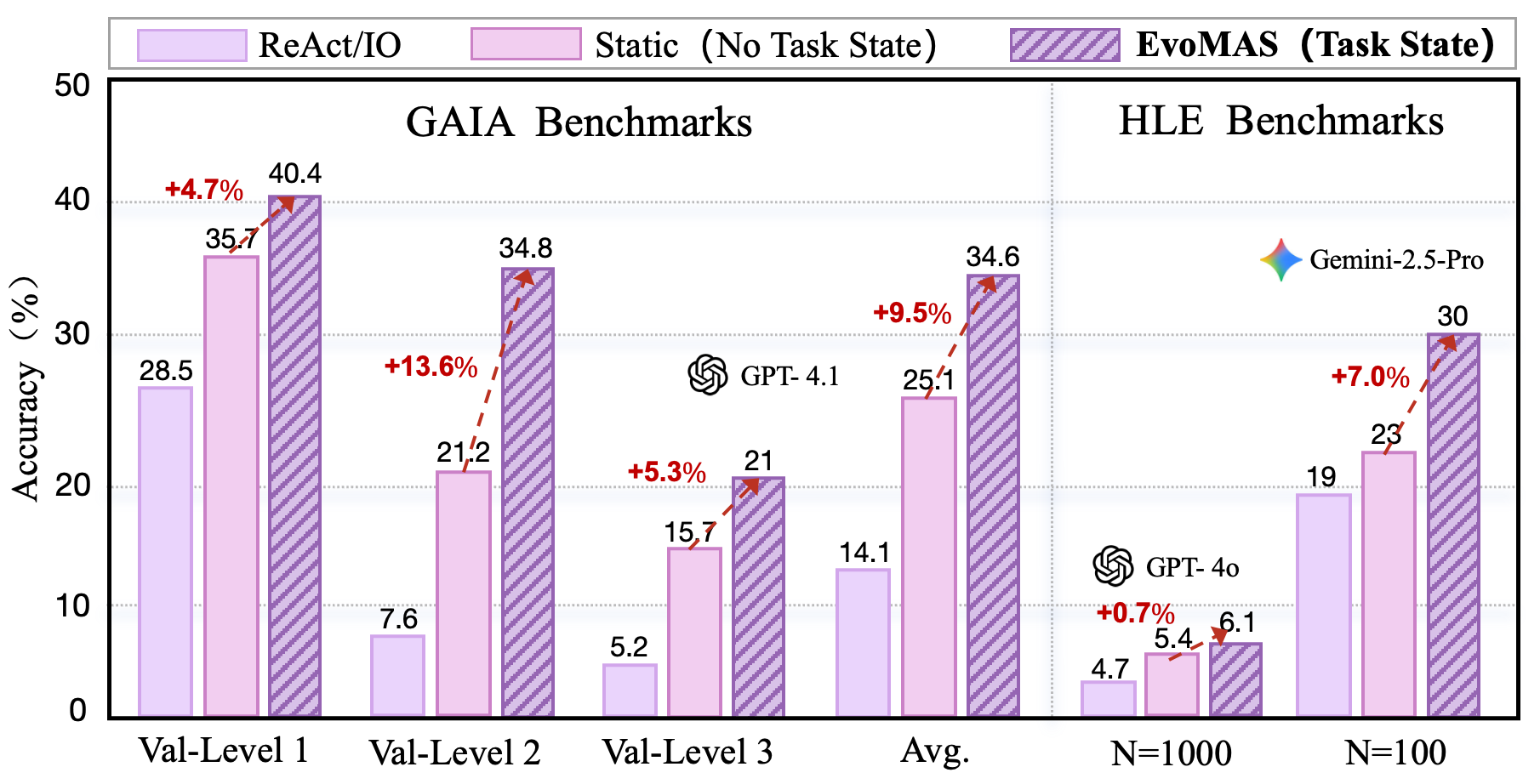}
    \caption{Effect of execution-time task state construction under a fixed workflow.}
    \label{fig:task_state}
\end{subfigure}
\hfill
\begin{subfigure}[t]{0.38\linewidth}
    \centering
    \includegraphics[width=\linewidth]{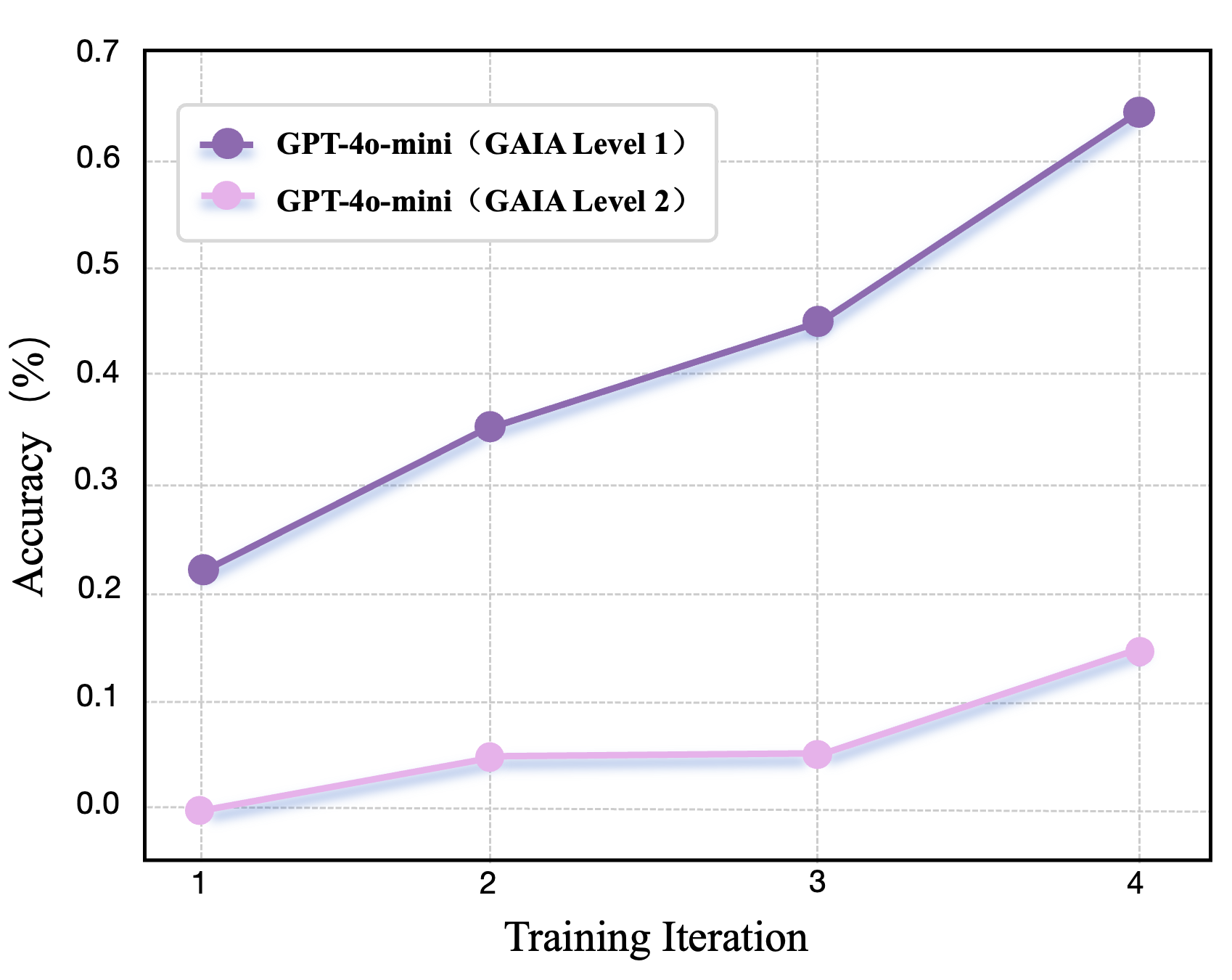}
    \caption{Training dynamics of the Workflow Adapter on GAIA.}
    \label{fig:training_curve}
\end{subfigure}
\caption{
\textbf{Task-state construction and workflow learning.}
(a) Explicit execution-time task-state construction improves success rates
under fixed workflows across GAIA and HLE.
(b) The Workflow Adapter learns steadily from sparse terminal reward on GAIA
Level 1 and Level 2.
}
\label{fig:state_and_learning}
\vspace{-1.5em}
\end{figure}

Figure~\ref{fig:task_state} shows that enabling task-state construction
consistently improves success rates on both benchmarks while leaving routing
and agent roles unchanged.
On GAIA, gains tend to be more pronounced on harder levels in the plotted
split, which aligns with the intuition that longer tasks benefit more from
structured bookkeeping of intermediate artifacts, partial failures, and
subgoal revisions.
On HLE, absolute accuracies remain challenging, but the lift from task-state
tracking is consistent across the two backbones, indicating that the effect is
not an artifact of a single model family.
Overall, these results isolate \emph{state construction} as a standalone
ingredient: EvoMAS improves not only from learned workflow selection, but also
from reliable execution-time state tracking.

\paragraph{Can the Workflow Adapter Learn from Sparse Terminal Reward?}
We next ask whether the Workflow Adapter can be optimized when supervision is
restricted to sparse, task-level terminal feedback.
We keep the 7-agent candidate pool and a 3-layer workflow template fixed, and
train only the adapter parameters with the REINFORCE-style objective described
in Section~\ref{subsec:optimization}; agent scoring follows the
cross-attention mechanism in Section~\ref{subsec:design}.
Figure~\ref{fig:training_curve} tracks held-out validation success on GAIA
Level~1 and Level~2 using GPT-4o-mini as the shared backbone.
Validation accuracy rises steadily on Level~1 and also improves on Level~2
throughout training, suggesting that terminal successes assign useful credit
to stage-wise workflow choices and that the learned policy transfers beyond the
easiest level used for early optimization.

Taken together, the curve supports using verifiable terminal reward as the
default training signal on moderate-difficulty regimes: even without dense
process supervision, the adapter learns a non-trivial state-conditioned
workflow policy rather than oscillating near initialization.

% Together, these two ablations separate the two sources of improvement in
% EvoMAS. The fixed-workflow experiment shows that maintaining an explicit
% execution-time task state is beneficial even without learning a new workflow,
% while the training curve shows that the Workflow Adapter can further exploit
% this state representation to learn better workflow choices. This supports our
% design of coupling task-state construction with state-conditioned workflow
% adaptation.

\begin{figure}[t]
  \centering
  \includegraphics[width=0.92\linewidth]{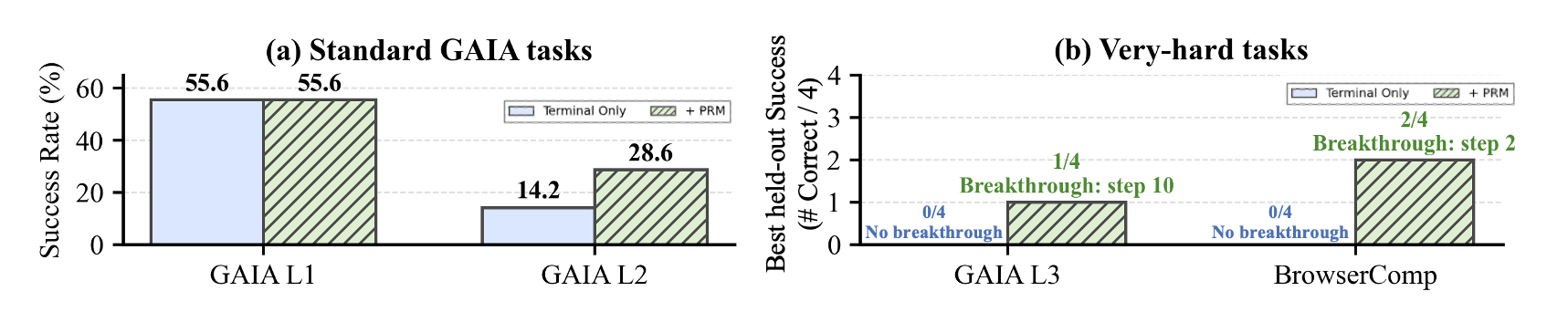}
  \caption{
  \textbf{Effect of process reward across task difficulty.} For standard GAIA tasks,
  bar height reports validation success rate. For very-hard tasks, bar height
  reports the best number of correct held-out validation tasks out of four; BT
  denotes the first training step with non-zero validation success. PRM is used
  in both training and inference.
  }
  \label{fig:prm}
  \vspace{-0.5em}
\end{figure}

\vspace{-0.7em}
\subsection{Additional Analyses}
\label{sec:additional_analysis}
% \vspace{-0.5em}

\paragraph{When Is Process Reward Necessary?}
We next study the role of evaluator-based process reward (PRM). The main
experiments use verifiable terminal reward only, since it is less dependent
on evaluator judgments. However, in very-hard settings, terminal reward can
be too sparse: if all sampled trajectories in a batch fail, policy-gradient
updates receive no positive learning signal. Figure~\ref{fig:prm} shows that
PRM is not universally required, but becomes useful under sparse-reward
regimes. On standard GAIA tasks, PRM leaves GAIA Level 1 unchanged and
improves GAIA Level 2 from 14.2 to 28.6. On very-hard tasks, PRM enables the
first non-zero validation breakthrough on GAIA Level 3 and BrowserComp, while
terminal-only training remains in the zero-success regime. Thus, PRM is best
viewed as a credit-assignment aid for extremely sparse terminal-reward
settings rather than as the default training signal.

% \begin{figure}[t]
% \centering
% \includegraphics[width=0.92\linewidth]{img/prm.png}
% \caption{
% Effect of process reward across task difficulty. For standard GAIA tasks,
% bar height reports validation success rate. For very-hard tasks, bar height
% reports the best number of correct held-out validation tasks out of four; BT
% denotes the first training step with non-zero validation success. PRM is used
% in both training and inference.
% }
% \label{fig:prm}
% \vspace{-0.5em}
% \end{figure}

% \vspace{0.4em}
\noindent
\begin{minipage}[t]{0.58\linewidth}
\paragraph{Sensitivity to Backbone Capability.}
We also test whether stronger Planner, Evaluator, and execution agents improve
EvoMAS while keeping the same workflow-adaptation framework. Replacing GPT-4o-mini with GPT-4o improves GAIA
Level 1 by 22.2 points and GAIA Level 2 by 7.2 points. This suggests that
EvoMAS is not tied to a specific lightweight backbone; under the same
workflow-adaptation framework, stronger underlying models can further improve
execution-time coordination.
\end{minipage}
\hfill
\begin{minipage}[t]{0.38\linewidth}
\centering
\captionof{table}{Backbone sensitivity on GAIA benchmarks.}
\label{tab:backbone}
\small
\setlength{\tabcolsep}{5pt}
\renewcommand{\arraystretch}{1.05}
\begin{tabular}{lcc}
\toprule
Backbone & L1 & L2 \\
\midrule
GPT-4o-mini & 55.6 & 14.2 \\
GPT-4o & \textbf{77.8} & \textbf{21.4} \\
$\Delta$ & +22.2 & +7.2 \\
\bottomrule
\end{tabular}
\end{minipage}
% \vspace{0.4em}

Overall, these analyses suggest that EvoMAS has different training-signal
requirements across difficulty regimes. Sparse terminal reward is sufficient
for standard GAIA tasks, while evaluator-based process reward becomes useful
when terminal success is too rare to provide learning signal. Meanwhile,
stronger backbones further improve performance, indicating that EvoMAS is
complementary to improvements in the underlying LLMs.

\subsection{Case Study: Workflow Adaptation Across Task States}
\label{subsec:case_study}
% \vspace{-0.4em}

\begin{figure}[t]
    \centering
    \includegraphics[width=\linewidth]{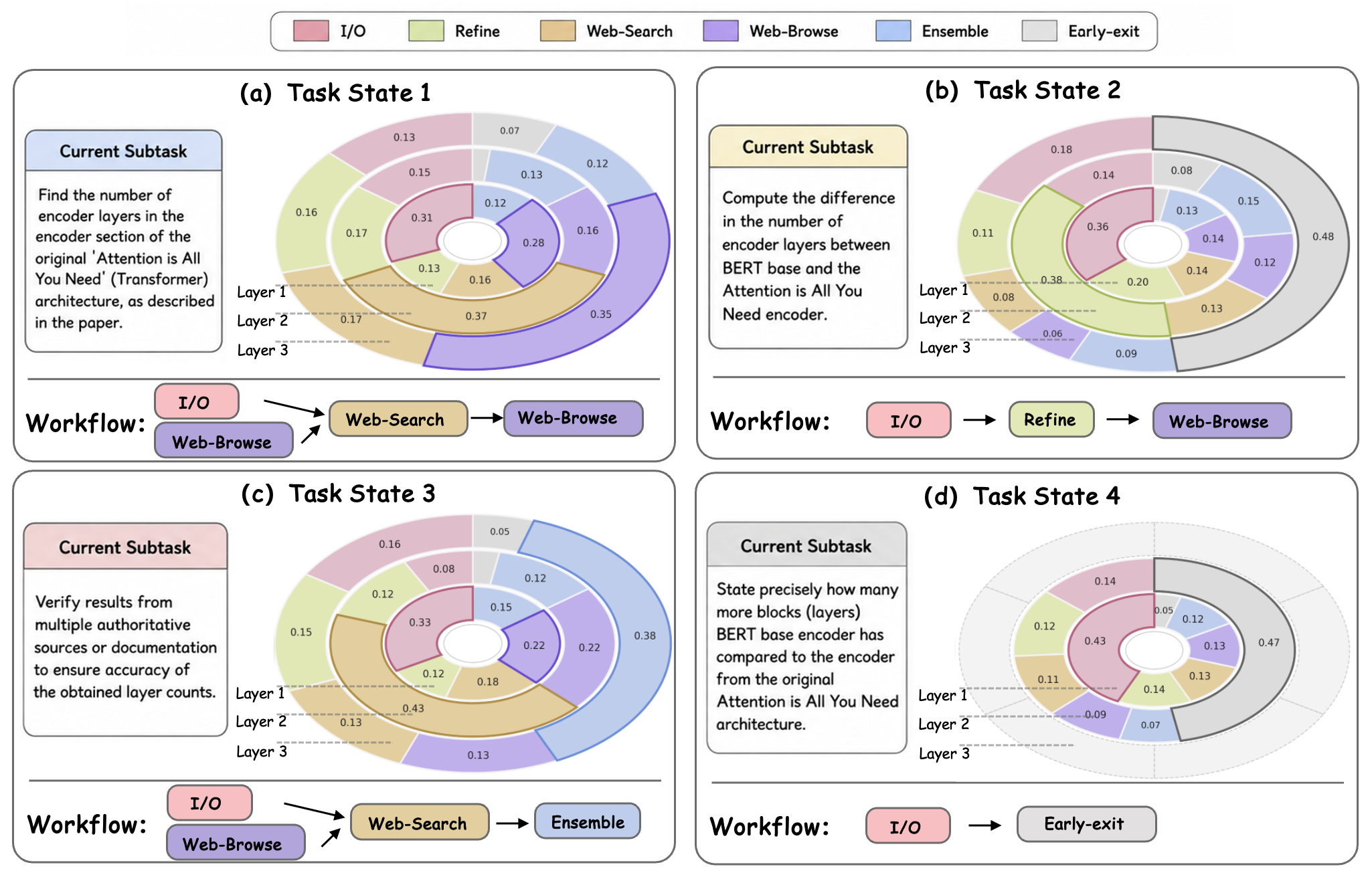}
    \caption{
    \textbf{Workflow Adapter behavior under different execution-time task states.}
    For the same agent pool, the adapter assigns different layer-wise selection
    probabilities and instantiates different workflows as the task state evolves.
    }
    \label{fig:adapter_behavior}
    % \vspace{-0.4em}
\end{figure}

\paragraph{State-dependent workflow choices.}
We first visualize what the Workflow Adapter learns across different
execution-time task states. Figure~\ref{fig:adapter_behavior} shows the
layer-wise selection probabilities under four task states from the same
trajectory. The candidate agent pool remains fixed, while each ring denotes a
workflow layer and each colored sector denotes an agent type. As the task
progresses, the adapter shifts probability mass toward different agents:
early states emphasize information acquisition and exploration, verification
states increase the preference for Web-search and Ensemble agents, and the
final state assigns high probability to Early-exit once sufficient evidence
has been accumulated.

\begin{figure}[t]
    \centering
    \includegraphics[width=\linewidth]{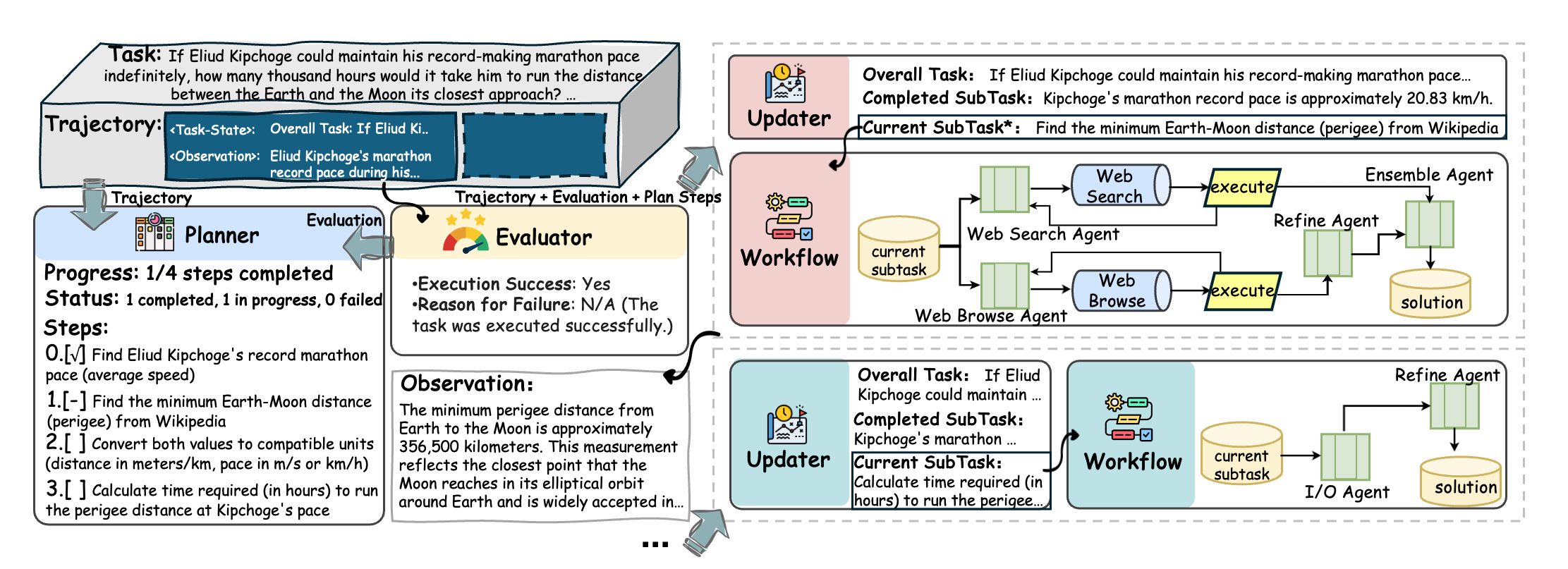}
    \caption{
    \textbf{End-to-end execution trace of EvoMAS on a GAIA task.}
    EvoMAS updates the task state across execution stages and constructs
    stage-specific workflows conditioned on the evolving state.
    }
    \label{fig:case_study}
    % \vspace{-0.4em}
\end{figure}

\paragraph{End-to-end execution trace.}
Figure~\ref{fig:case_study} further illustrates how task-state construction
and workflow adaptation interact during execution. On a GAIA task about
computing how long it would take Eliud Kipchoge to run the closest
Earth--Moon distance at his marathon pace, EvoMAS decomposes the task into
subtasks, executes a stage-specific workflow, evaluates the intermediate
result, and updates the task state. The updated state then conditions the next
workflow, enabling EvoMAS to switch from retrieving Kipchoge's pace, to
finding the Earth--Moon perigee distance, to computing the final answer. This
case shows that EvoMAS adapts coordination structures throughout the task
rather than applying a single static workflow.

% \vspace{-1em}
\section{Related Work}
% \vspace{-0.3em}
\paragraph{LLM-Agents and Agentic Systems.}
Recent work has explored equipping large language models with tools, memory, and
iterative reasoning mechanisms to build autonomous agents capable of closed-loop
task execution~\citep{react,toolformer,hugginggpt,reflexion,memgpt,voyager}.
Beyond single-agent settings, multi-agent systems (MAS) further improve performance
through role specialization, debate or critique mechanisms, and structured coordination,
as demonstrated in frameworks such as AutoGen, LLM-Debate, and AgentVerse~\citep{autogen,llmdebate,agentverse}.
While these agentic systems substantially extend the capabilities of individual models,
their coordination structures and interaction patterns are typically manually specified,
resulting in fixed agent roles and static workflows~\citep{autogen,metagpt,NeurIPS2023camel,wang2023agentsurvey}.
Such hand-crafted designs limit the flexibility of agentic systems when applied to
complex tasks requiring sustained interaction and iterative refinement.
\vspace{-1em}

\paragraph{Automating Multi-Agent Systems.}
Recent work has sought to automate the design of multi-agent systems by reducing
manual specification of prompts, roles, and coordination structures.
Representative approaches include prompt and module optimization,
automated role or profile construction, and learning or searching over
inter-agent communication patterns and workflows~\citep{promptbreeder,dspy,evoprompt,evoagent,agentprune,gptswarm,aflow,a2flow,gdesigner,maas,metaagent,evomac}.
These methods demonstrate that automated system design can outperform
hand-crafted agent configurations on a range of tasks~\citep{gptswarm,aflow,gdesigner,agentprune,maas}.
However, most existing approaches still follow a \emph{one-shot} paradigm,
optimizing a single agentic system or workflow prior to execution and reusing it
unchanged throughout the task, resulting in static workflows without
execution-time adaptation~\citep{gptswarm,aflow,gdesigner,maas,agentprune}.

\vspace{-0.5em}
\section{Conclusion}
\vspace{-0.5em}
% This paper introduces EvoMAS, a framework for execution-time multi-agent
% workflow construction that couples structured task-state tracking with a
% state-conditioned policy over layered workflows, moving beyond one-shot,
% initialization-time coordination designs. On GAIA, HLE, and DeepResearcher,
% EvoMAS outperforms strong single-agent baselines and recent automated MAS
% methods. Ablations show complementary gains from explicit task-state
% construction and learned workflow adaptation; additional analyses highlight when
% process reward and stronger backbones matter most.

% EvoMAS still uses a fixed candidate pool and maximum workflow depth; scaling
% the pool, open-ended agent creation, and runtime depth adaptation are important
% next steps. Richer credit assignment may further help very-hard regimes beyond
% our PRM-focused study.

This paper introduces EvoMAS, a framework for execution-time multi-agent
workflow construction that combines structured task-state tracking with
state-conditioned layered workflow adaptation. On GAIA, HLE, and
DeepResearcher, EvoMAS outperforms strong single-agent baselines and recent
automated MAS methods. Our analyses show complementary gains from explicit
task-state construction and learned workflow adaptation, and identify when
process reward and stronger backbones are most useful.

EvoMAS still assumes a fixed candidate pool and maximum workflow depth.
Scaling to larger pools, open-ended agent creation, runtime depth adaptation,
and richer credit assignment remain important future directions.

\newpage
% TEMP for regenerating .bbl locally (will be switched back to \input{...bbl})
\bibliographystyle{plainnat}
\bibliography{example_paper}

\appendix

% ============================================================
\section{Algorithmic Summary of EvoMAS}
\label{sec:appendix_algorithm}

This section provides a high-level pseudocode description of EvoMAS,
summarizing the interaction between execution-time task-state construction,
state-conditioned workflow adaptation, and policy optimization.
Algorithm~\ref{alg:evomas} corresponds to the meta-level formulation in
Section~\ref{sec:formal} and the method description in Section~\ref{sec:method}.

% Optional: make comments compact and aligned.
\algrenewcommand\algorithmiccomment[1]{\hfill{\footnotesize$\triangleright$ #1}}

\begin{algorithm}[t]
\caption{EvoMAS: Execution-Time Workflow Construction and Learning}
\label{alg:evomas}
\small
\begin{algorithmic}[1]
\Require Task instance $\tau$; fixed agent pool $\mathcal{P}$; workflow policy
$\pi_{\mathrm{wf}}(\cdot \mid \tilde{s};\Theta)$; maximum meta-steps $T$
\Ensure Final answer $\hat{a}(\tau)$; updated policy parameters $\Theta$ if training

\State Initialize structured task state $\tilde{s}_0$ from $\tau$
\Comment{initial plan and active subtask}
\State Initialize trajectory buffer $\mathcal{D}\leftarrow \emptyset$
\Comment{stores sampled decisions}

\For{$t=1$ \textbf{to} $T$}
    \State \textit{// Construct a stage-specific workflow}
    \State Sample workflow $\mathcal{G}_t \sim
    \pi_{\mathrm{wf}}(\cdot \mid \tilde{s}_{t-1};\Theta)$
    \Comment{state-conditioned adaptation}
    \State Store $(\tilde{s}_{t-1},\mathcal{G}_t)$ in $\mathcal{D}$
    \Comment{for policy-gradient update}

    \State \textit{// Execute the workflow}
    \State Execute $\mathcal{G}_t$ conditioned on $\tilde{s}_{t-1}$
    \State Obtain aggregated execution record $\xi_t$
    \Comment{outcomes, messages, tool observations, artifacts}

    \State \textit{// Evaluate and update the task state}
    \State $e_t \leftarrow
    \mathrm{Evaluator}(\tau,\tilde{s}_{t-1},\xi_t)$
    \Comment{verdict, confidence, feedback}
    \State $\xi_t \leftarrow (\xi_t, e_t)$
    \Comment{include assessment signals in $\xi_t$}
    \State $\tilde{s}_t \leftarrow
    \mu(\tau,\tilde{s}_{t-1},\mathcal{G}_t,\xi_t)$
    \Comment{Planner--Evaluator--Updater transition}

    \If{task is completed \textbf{or} Early-exit is selected}
        \State \textbf{break}
    \EndIf
\EndFor

\State Extract final answer $\hat{a}(\tau)$ from $\tilde{s}_t$ and/or $\xi_t$

\If{training}
    \State Compute terminal utility $U(\tau)$ using the benchmark evaluator
    \Comment{verifiable terminal reward}
    \State Update $\Theta$ using REINFORCE with trajectory buffer $\mathcal{D}$
    \Comment{trajectory-level return}
\EndIf

\State \Return $\hat{a}(\tau)$
\end{algorithmic}
\end{algorithm}

During training, the terminal utility is used as the trajectory-level return.
The policy-gradient update is
\begin{equation}
\nabla_\Theta \mathcal{J}(\Theta)
\approx
U(\tau)
\sum_{(\tilde{s},\mathcal{G})\in\mathcal{D}}
\nabla_\Theta \log P_\Theta(\mathcal{G}\mid \tilde{s}).
\end{equation}

\section{Task State Construction Details}
\label{app:state_construction}

\subsection{Planner Details}
\label{app:planner_details}

The Planner maintains an ordered subtask plan
\begin{equation}
P_\tau=\langle u_1,\ldots,u_H\rangle,
\end{equation}
where each subtask $u_h$ is associated with a status
\begin{equation}
\sigma(u_h)\in\{\texttt{pending},\texttt{completed},\texttt{failed}\}.
\end{equation}
At each meta-step, the Planner updates the active subtask according to the
previous execution outcome and evaluator feedback. If the active subtask is
successfully solved, it is marked as \texttt{completed} and the next pending
subtask becomes active. If execution fails or evidence is insufficient, the
Planner may keep the same subtask and attach failure context for refinement.
When repeated failures or unexpected constraints are observed, the Planner may
revise the subtask plan by adding, removing, or reordering subtasks.

\subsection{Evaluator Details}
\label{app:evaluator_details}

Given the previous task state $\tilde{s}_{t-1}$ and execution outcome
$\xi_{t-1}$, the Evaluator produces a structured assessment
\begin{equation}
e_{t-1} =
(\mathrm{verdict}_{t-1}, c_{t-1}, f_{t-1}),
\end{equation}
where $\mathrm{verdict}_{t-1}\in
\{\texttt{success},\texttt{fail},\texttt{refine}\}$ indicates whether the active
subtask is solved, failed, or requires further refinement, $c_{t-1}\in[0,1]$
is a confidence score, and $f_{t-1}$ is natural-language feedback describing
detected issues, missing evidence, or suggested corrections. These signals are
used by the Updater to construct the next task state, and are included in the
aggregated execution record $\xi_{t-1}$ in the Meta-MDP notation.

\subsection{Updater and State Serialization}
\label{app:updater_details}

The Updater implements the meta-level transition by integrating the previous
task state, the execution outcome, and evaluator feedback. The serialized task
state contains the following fields:
\begin{itemize}
    \item overall task objective;
    \item current active subtask;
    \item completed subtasks and their intermediate artifacts;
    \item evaluator verdict, confidence, and feedback;
    \item failure or refinement context when available;
    \item compact execution history relevant to future workflow decisions.
\end{itemize}
This serialization is used in two ways. A compact version centered on the
active subtask is used to condition the Workflow Adapter, while the full
structured state is injected into instantiated agents as execution-time
working context.

\section{Prompt Templates}
\label{app:prompts}

\paragraph{Implementation mapping.}
In our implementation, the Planner is realized by the PlanningAgent together
with the planning tool, the Evaluator is implemented by the JudgeAgent, and
the Updater is implemented as deterministic state aggregation and
serialization logic inside the executor tool. Thus, the Updater in the paper
denotes the state-construction module that integrates the plan state,
execution outcome, and evaluator feedback, rather than a separate LLM agent.

\subsection{Planner Prompt}
\label{app:planner_prompt}

The Planner updates the subtask plan based on the previous task state, the
latest execution outcome, and evaluator feedback. In the implementation, this
role is realized by the PlanningAgent (\texttt{plan\_agent/prompts/planning\_agent.yaml})
and the planning tool (\texttt{tools/planning.py}).

The following template summarizes the prompt protocol used by the Planner:

\begin{promptbox}{Prompt 1. \textsc{Planner} template for execution-time subtask planning.}
\begin{PromptVerbatim}
You are the Planner in an execution-time multi-agent system.
Your job is to update the subtask plan based on the previous
task state, the latest execution outcome, and the evaluator
feedback.

[INPUT]
Overall task: {{overall_task}}
Previous task state: {{prev_task_state}}
Latest execution outcome: {{execution_outcome}}
Evaluator assessment: {{evaluator_feedback}}

[OUTPUT]
1) Updated subtask list with status tags:
   [TODO / DONE / FAILED / NEED-REFINE]
2) Next active subtask: one sentence
3) Updated completed-subtasks summary: concise and
   reusable in later stages

[RULES]
- Mark a subtask DONE only when evaluator feedback
  indicates success.
- If execution failed or evidence is missing, propose
  a repair/refinement subtask.
- Keep the plan minimal and focused on the next
  execution stage.
\end{PromptVerbatim}
\end{promptbox}

\subsection{Evaluator Prompt}
\label{app:evaluator_prompt}

The Evaluator is implemented as a tool-augmented JudgeAgent that verifies whether
the current active subtask has been successfully completed. Our implementation
(\texttt{judge\_agent/prompts/judge\_agent.yaml}) equips the Evaluator with powerful
verification tools including deep research capabilities and systematic analysis tools,
enabling it to conduct thorough validation beyond simple pattern matching.

The JudgeAgent has access to:
\begin{itemize}[nosep,leftmargin=1em]
\item \texttt{final\_judge\_answer\_tool}: Returns structured feedback and binary reward
\item \texttt{deep\_researcher\_tool}: Conducts extensive web searches for fact verification
\item \texttt{deep\_analyzer\_tool}: Performs systematic, step-by-step analysis
\end{itemize}

This tool-augmented design allows the Evaluator to actively verify claims, cross-check
facts, and perform deep analysis when assessing execution outcomes, rather than relying
solely on surface-level heuristics. The final output includes natural-language feedback
and a binary reward. For consistency with the paper-level abstraction, we map reward 1
to a successful verdict and reward 0 to a failed or refinement-needed verdict, depending
on the feedback.

\begin{promptbox}{Prompt 2. \textsc{Evaluator} template for step verification.}
\begin{PromptVerbatim}
You are the Evaluator. Determine whether the CURRENT
subtask is completed given the execution outcome.

You have access to powerful verification tools:
- deep_researcher_tool: conduct web searches to verify
  facts and claims
- deep_analyzer_tool: perform systematic analysis to
  validate reasoning and completeness
- final_judge_answer_tool: provide structured feedback
  and binary reward

[INPUT]
Overall task: {{overall_task}}
Current active subtask: {{current_step_description}}
Execution outcome: {{execution_result}}

[OUTPUT]
Use final_judge_answer_tool to provide:
- feedback: detailed diagnosis of what was accomplished,
  what is missing, or what is incorrect
- reward: binary score (0 for failure, 1 for success)

[RULES]
- Evaluate ONLY the current active subtask, not the
  entire overall task.
- Use verification tools actively when claims need
  fact-checking or when reasoning needs validation.
- If the outcome is partially correct but incomplete,
  provide reward=0 with constructive feedback.
- If the outcome sufficiently solves the current
  subtask, provide reward=1.
\end{PromptVerbatim}
\end{promptbox}

\subsection{Updater State Serialization}
\label{app:updater_prompt}

The Updater integrates the plan state, evaluator assessment, execution
outcome, and compact execution history into the next structured task state.
In the implementation, this step is performed by deterministic aggregation
and serialization logic inside the executor tool (\texttt{tools/executor\_tool.py})
rather than by a separate LLM agent.

\begin{promptbox}{Prompt 3. \textsc{Updater} state-serialization template.}
\begin{PromptVerbatim}
You are the Updater. Integrate planning state, evaluator
assessment, and execution outcome into the next structured
task state.

[INPUT]
Plan state: {{plan_state}}
Evaluator assessment: {{evaluator_assessment}}
Execution outcome: {{execution_outcome}}
Compact execution history: {{history}}

[OUTPUT: STRUCTURED TASK STATE]
- overall_objective: ...
- active_subtask: ...
- completed_subtasks: bullet list with key
  artifacts/results
- issues_and_feedback: ...
- next_stage_hint: what information, tools, or agents
  may be needed next

[RULES]
- Preserve useful intermediate artifacts.
- Keep the state compact enough for workflow selection.
- Include failure or refinement context when evaluator
  feedback indicates it.
\end{PromptVerbatim}
\end{promptbox}

\subsection{Candidate Agent Prompt Templates}
\label{app:agent_prompts}

For readability, we present lightly normalized prompt templates that preserve
the role, input fields, and output constraints used in the implementation.

Table~\ref{tab:agent_prompt_index} lists the core prompt files for the
candidate agents in our pools. Paths follow the released codebase layout
(\texttt{atomic\_agents/} is the implementation module name); agent names match
the main text.

\begin{table}[t]
\centering
\caption{Core candidate-agent prompt template paths.}
\label{tab:agent_prompt_index}
\small
\setlength{\tabcolsep}{7pt}
\renewcommand{\arraystretch}{1.18}
\begin{tabularx}{\textwidth}{@{}>{\raggedright\arraybackslash}p{2.05cm}>{\raggedright\arraybackslash}p{2.25cm}>{\raggedright\arraybackslash}X@{}}
\toprule
\textbf{Agent} & \textbf{Role} & \textbf{Implementation prompt file} \\
\midrule
I/O & Direct solving & \url{atomic_agents/io_agent/prompts/io_agent.yaml} \\
Multi-generate & Diverse generation & \url{atomic_agents/multi_generate_agent/prompts/multi_generate_agent.yaml} \\
Self-refine & Refinement & \url{atomic_agents/self_refine_agent/prompts/self_refine_agent.yaml} \\
Web-search & Web retrieval & \url{atomic_agents/web_search_agent/prompts/web_search_agent.yaml} \\
Web-browser & Web browsing & \url{atomic_agents/web_browser_agent/prompts/web_browser_agent.yaml} \\
Early-exit & Termination & \url{atomic_agents/early_exit_agent/prompts/early_exit_agent.yaml} \\
Ensemble & Aggregation & \url{atomic_agents/sc_ensemble_agent/prompts/sc_ensemble_agent.yaml} \\
\bottomrule
\end{tabularx}
\end{table}

The implementation also includes auxiliary agents for query generation, file
reading, code execution, and generic tool use. These are implementation
helpers and are not part of the default 4-agent or 7-agent pools reported in
the main experiments.

\subsubsection{I/O Agent}
The I/O agent directly solves the current subtask using internal knowledge.

\begin{promptbox}{I/O Agent prompt template.}
\begin{PromptVerbatim}
You are a knowledge-based assistant that answers
questions using your internal knowledge.

[TASK]
{{problem}}

[INSTRUCTION]
Please answer this question using your internal
knowledge. You may use the provided task context if
helpful.

At the end of your response, you must add a
clarification statement indicating that the answer
comes from the model's internal knowledge base.
\end{PromptVerbatim}
\end{promptbox}

\subsubsection{Multi-generate Agent}
The Multi-generate agent produces diverse candidate solutions for the current task.

\begin{promptbox}{Multi-generate Agent prompt template.}
\begin{PromptVerbatim}
You are a knowledge-based assistant that answers
questions using your internal knowledge.

[TASK]
{{task}}

[INSTRUCTION]
Please answer this question using your internal
knowledge. You may refer to the solution above if
helpful.

Important Guidelines:
- If the current task contains explicit keywords
  related to "verification" or "validation", you
  should compare all possible answers to verify
  whether previous answers are correct - do not
  blindly trust the answers
\end{PromptVerbatim}
\end{promptbox}

\subsubsection{Self-refine Agent}
The Self-refine agent analyzes solutions against task requirements, identifies errors, and provides refined versions.

\begin{promptbox}{Self-refine Agent prompt template.}
\begin{PromptVerbatim}
You are a solution refinement assistant. Your role
is to analyze solutions against the given task,
identify errors and missing details, and provide a
corrected and improved version.

[PROBLEM]
{{problem}}

[CURRENT SOLUTION]
{{solution}}

[INSTRUCTION]
Your Process:
1. Analyze: Carefully examine the current solution
   in the context of the task requirements.
2. Evaluate: Identify factual errors, missing
   information, incomplete aspects, logical
   inconsistencies.
3. Refine: Correct all identified errors and
   supplement missing details.

Important Guidelines:
- If the task has multiple possible solutions,
  carefully evaluate whether the provided current
  solution is correct - do not blindly trust it
- Base your analysis strictly on the task
  requirements
- Correct factual errors and logical mistakes
- Add missing details necessary to fully answer
  the task
\end{PromptVerbatim}
\end{promptbox}

\subsubsection{Web-search Agent}
The Web-search agent searches the web for relevant information and provides answers with source citations.

\begin{promptbox}{Web-search Agent prompt template.}
\begin{PromptVerbatim}
You are a web search specialist assistant. Your
role is to answer questions by searching the web
for relevant information, then providing clear and
accurate answers.

You have access to web_searcher_tool to search for
information relevant to the task. Once you have
enough information, use final_answer_tool to
provide your answer.

[TASK]
Search the web to find information and answer the
following question:
{{task}}

[RULES]
1. ALWAYS call a tool - use web_searcher_tool to
   search or final_answer_tool to answer
2. Use clear and specific search queries to get
   relevant results
3. Never repeat the same search query
4. Always provide your final answer using
   final_answer_tool
5. Include source URLs in your final answer - add
   a "Sources:" section with URLs from the search
   results you used
\end{PromptVerbatim}
\end{promptbox}

\subsubsection{Web-browser Agent}
The Web-browser agent interacts with web pages, extracts information, and answers questions by browsing websites.

\begin{promptbox}{Web-browser Agent prompt template.}
\begin{PromptVerbatim}
You are a web browser automation specialist
assistant. Your role is to interact with web pages,
extract information, and answer questions by
browsing websites and performing various browser
actions.

You have access to custom_browser_tool to interact
with web pages - navigate to URLs, click elements,
input text, scroll, extract content, etc.

[TASK]
Browse the web and interact with web pages to
answer the following question:
{{task}}

[RULES]
1. ALWAYS call a tool - use custom_browser_tool to
   browse and interact with web pages or
   final_answer_tool to answer
2. Navigate to relevant websites first, then
   browse and interact with web pages or
   final_answer_tool to answer
2. Navigate to relevant websites first, then
   interact with elements (click, input, scroll)
   as needed
3. Use extract_content action to get specific
   information from the current page
4. Always provide your final answer using
   final_answer_tool
5. Include source URLs in your final answer - add
   a "Sources:" section with URLs from the web
   pages you visited
\end{PromptVerbatim}
\end{promptbox}

\subsubsection{Early-exit Agent}
The Early-exit agent serves as a signal to terminate the multi-layer agent coordination process early.

\begin{promptbox}{Early-exit Agent prompt template.}
\begin{PromptVerbatim}
You are an early exit signal agent. When selected
by the Coordinator, you indicate that the current
task execution should stop immediately.

This agent is used as a signal to terminate the
multi-layer agent coordination process early.

You do not need to perform any actual task
execution.
\end{PromptVerbatim}
\end{promptbox}

\subsubsection{Ensemble Agent}
The Ensemble agent evaluates multiple solutions and identifies the most consistent answer across them.

\begin{promptbox}{Ensemble Agent prompt template.}
\begin{PromptVerbatim}
You are a solution evaluation specialist. Your role
is to carefully evaluate multiple solutions for the
CURRENT TASK (not the overall task) and identify
the answer that appears most frequently across
them. This consistency in answers is crucial for
determining the most reliable solution.

[PROBLEM]
Given the question described as follows:
{{problem}}

[SOLUTIONS]
Several solutions have been generated to address
the CURRENT TASK above. They are as follows:
{{solutions}}

[INSTRUCTION]
Your role is to evaluate these solutions for the
CURRENT TASK and identify the answer that appears
most frequently across them. This consistency in
answers is crucial for determining the most
reliable solution for the current task.

In the "thought" field, provide a detailed
explanation of your thought process, focusing on
which solution best addresses the current task. In
the "solution_letter" field, output only the single
letter ID (A, B, C, etc.) corresponding to the most
consistent solution for the current task. Do not
include any additional text or explanation in the
"solution_letter" field.
\end{PromptVerbatim}
\end{promptbox}

\section{Implementation Details}

\subsection{Workflow Adapter Details}
\label{app:workflow_adapter}

Each candidate agent is represented by an embedding derived from its textual
role specification, capabilities, and tool bindings. We stack agent embeddings
as
\begin{equation}
\mathbf{E}=[\mathbf{e}_1;\ldots;\mathbf{e}_N]\in\mathbb{R}^{N\times d}.
\end{equation}
The compact task-state representation is encoded as
$\mathbf{X}_t\in\mathbb{R}^{M\times d}$. To incorporate cross-layer context, we
summarize previously selected agent embeddings at layer $\ell$ as
\begin{equation}
\mathbf{c}_t^{(\ell)}
=
\operatorname{Pool}
\bigl(
\{\mathbf{e}_i \mid a_i \in \mathcal{V}_t^{(<\ell)}\}
\bigr),
\end{equation}
where mean pooling is used in our implementation and
$\mathbf{c}_t^{(1)}=\mathbf{0}$. This context vector is appended to the task
state representation:
\begin{equation}
\tilde{\mathbf{X}}_t^{(\ell)}
=
[\mathbf{X}_t;\mathbf{c}_t^{(\ell)}].
\end{equation}

The Workflow Adapter applies cross-attention between task-state tokens and
agent embeddings:
\begin{equation}
\mathbf{Q}_t^{(\ell)}=\tilde{\mathbf{X}}_t^{(\ell)}\mathbf{W}_Q,
\qquad
\mathbf{K}=\mathbf{E}\mathbf{W}_K,
\qquad
\mathbf{V}=\mathbf{E}\mathbf{W}_V,
\end{equation}
and obtains
\begin{equation}
\mathbf{H}_t^{(\ell)}
=
\operatorname{Attn}
(\mathbf{Q}_t^{(\ell)},\mathbf{K},\mathbf{V}).
\end{equation}
The layer-specific query vector is obtained by mean pooling:
\begin{equation}
\mathbf{h}_t^{(\ell)}
=
\operatorname{Pool}(\mathbf{H}_t^{(\ell)}).
\end{equation}
The main text then uses $\mathbf{h}_t^{(\ell)}$ to score candidate agents and
sample a layer-wise agent subset.

\paragraph{Cumulative-mass soft sampling log-probability.}
For completeness, we detail the log-probability term used by REINFORCE under
our implementation of cumulative-mass soft sampling (Algorithm
\ref{alg:evomas}, line~\ref{alg:evomas}). At layer $\ell$, we form a categorical
distribution over agents,
\begin{equation}
\mathbf{p}_{t}^{(\ell)}
=
\mathrm{softmax}\!\left(\boldsymbol{\alpha}_{t}^{(\ell)}/\lambda\right),
\qquad
p_{t,i}^{(\ell)}
=
\frac{\exp(\alpha_{t,i}^{(\ell)}/\lambda)}{\sum_{j=1}^{N}\exp(\alpha_{t,j}^{(\ell)}/\lambda)},
\end{equation}
and then sample \emph{without replacement} until the cumulative selected
probability mass reaches the threshold $\rho$. Let
$\mathbf{i}_t^{(\ell)}=(i_{t,1}^{(\ell)},\ldots,i_{t,K_\ell}^{(\ell)})$ denote
the sampled index sequence at layer $\ell$, where $K_\ell$ is the (random)
stopping time induced by $\rho$. The resulting ordered-sequence probability is
\begin{equation}
P_\Theta(\mathbf{i}_t^{(\ell)}\mid\tilde{s}_t)
=
\prod_{r=1}^{K_\ell}
\frac{p_{t,i_{t,r}^{(\ell)}}^{(\ell)}}{1-\sum_{q<r}p_{t,i_{t,q}^{(\ell)}}^{(\ell)}},
\end{equation}
with the stopping condition
\begin{equation}
\sum_{r=1}^{K_\ell-1}p_{t,i_{t,r}^{(\ell)}}^{(\ell)}<\rho,
\qquad
\sum_{r=1}^{K_\ell}p_{t,i_{t,r}^{(\ell)}}^{(\ell)}\ge\rho.
\end{equation}
Our implementation further applies a fallback to selecting the
highest-probability agent when no agent is selected due to numerical or
degenerate edge cases.

\subsection{Agent Pool Specification}
Table~\ref{tab:agent_pools} lists the candidate agent pools used by EvoMAS.
The 4-agent pool provides basic capabilities for information retrieval and aggregation,
while the 7-agent pool adds multi-generation, self-refinement, and web-browsing capabilities.

\begin{table}[h]
\centering
\caption{Candidate agent pools used by EvoMAS.}
\label{tab:agent_pools}
\small
\begin{tabular}{ll}
\toprule
Pool & Agents \\
\midrule
4-agent & I/O, Early-exit, Web-search, Ensemble \\
7-agent & I/O, Multi-generate, Self-refine, Web-search, \\
        & Web-browser, Early-exit, Ensemble \\
\bottomrule
\end{tabular}
\end{table}

\subsection{Workflow Space and Edge Construction}
Given the selected agent sets $\{V_t^{(\ell)}\}_{\ell=1}^{L_t}$, we construct
inter-layer communication edges as fully connected bipartite connections.
The sampling procedure is applied separately at each layer, so the same agent
may appear in multiple layers; this allows repeated refinement. Across
meta-steps, agents may also be selected repeatedly.
Each downstream agent receives an aggregated message containing all upstream outputs, and
agent-specific prompts determine how the aggregated input is consumed.

\section{Experimental Details}
\label{app:experimental_details}

\subsection{Dataset Sampling and Splits}
Table~\ref{tab:data_splits} summarizes how each benchmark defines its task pool
and the train versus held-out evaluation partitions. Unless otherwise specified,
reported metrics follow the evaluation partition listed here.

For GAIA Level~1 and Level~2, we use the full released task set at each level and
apply a random 8:2 train--validation split (no additional subsampling).
For HLE and DeepResearcher, we draw 50 instances and split them 8:2.
GAIA Level~3 and BrowserComp are used only for very-hard sparse-reward analysis,
each with 20 sampled instances and a 16/4 train--validation split.

\begin{table}[t]
\centering
\caption{Dataset sampling and splits.}
\label{tab:data_splits}
\small
\setlength{\tabcolsep}{6pt}
\renewcommand{\arraystretch}{1.08}
\begin{tabular}{lccc}
\toprule
\textbf{Benchmark} & \textbf{Task pool} & \textbf{Train} & \textbf{Evaluation} \\
\midrule
GAIA Level 1 & Benchmark subset & Train split & Held-out evaluation split \\
GAIA Level 2 & Benchmark subset & Train split & Held-out evaluation split \\
GAIA Level 3 & 20 sampled instances & 16 & 4 \\
HLE & 50 sampled instances & 40 & 10 \\
DeepResearcher & 50 sampled instances & 40 & 10 \\
BrowserComp & 20 sampled instances & 16 & 4 \\
\bottomrule
\end{tabular}
\end{table}

\subsection{Model Backbones}
Table~\ref{tab:model_identifiers} lists the model identifiers used for different
components in our experiments.

\begin{table}[h]
\centering
\caption{Model identifiers used in our experiments.}
\label{tab:model_identifiers}
\small
\begin{tabular}{ll}
\toprule
Component & Model Identifier \\
\midrule
Planner & GPT-4o-mini / GPT-4o \\
Evaluator & GPT-4o-mini / GPT-4o \\
Updater & Deterministic executor-tool serialization \\
Execution agents & GPT-4o-mini / GPT-4o \\
Baselines & GPT-4o-mini / GPT-4o \\
\bottomrule
\end{tabular}
\end{table}

\subsection{Baseline Implementation Details}
GPTSwarm, AFlow, G-Designer, and MaAS are implemented following their
original papers. All baselines use the same backbone models and execution
budgets as EvoMAS for fair comparison.

\section{Additional Workflow Adapter Visualizations}
\label{app:adapter_visualization}

\begin{figure}[t]
    \centering
    \includegraphics[width=0.95\linewidth]{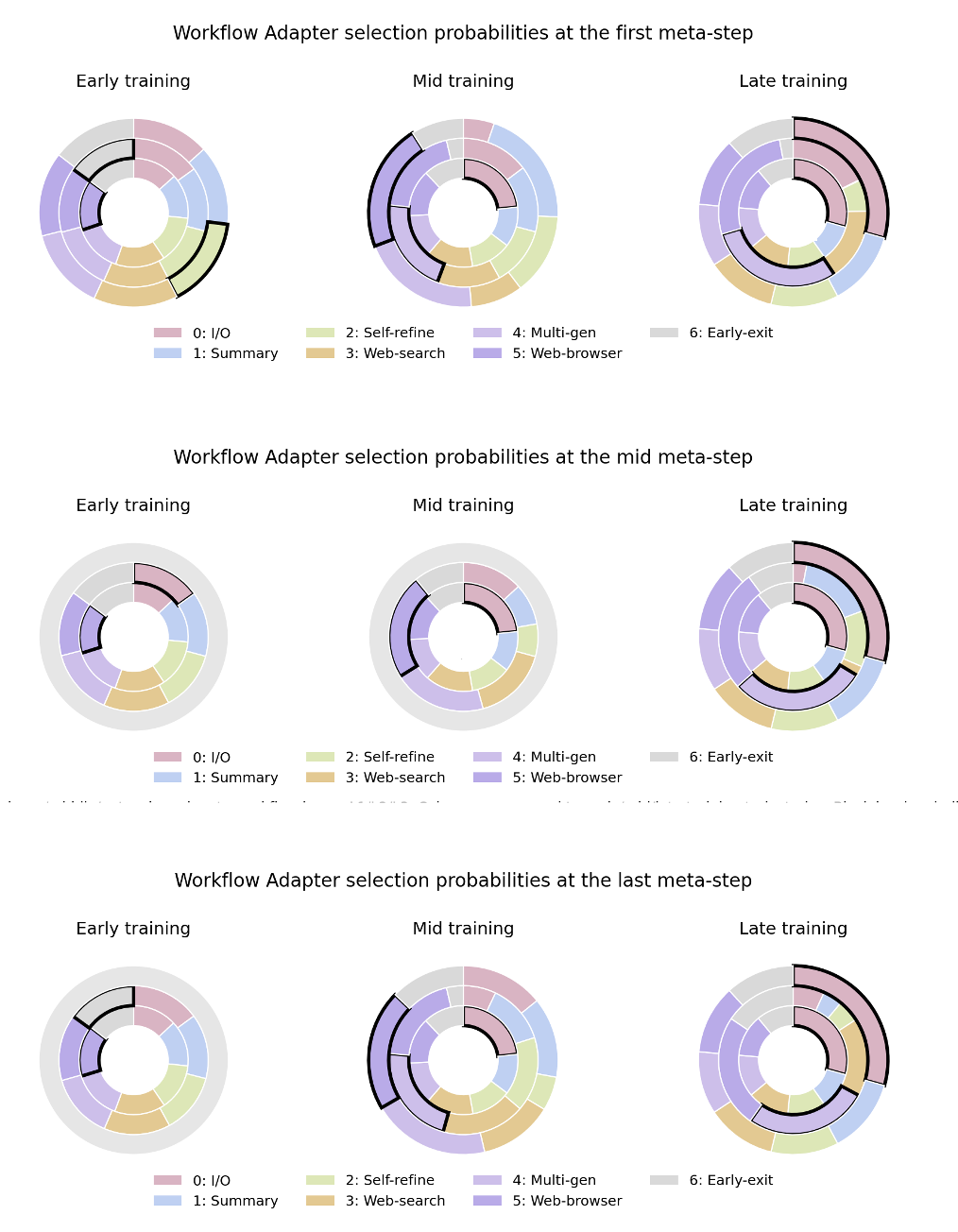}
    \caption{
    Validation-time evolution of Workflow Adapter selection probabilities.
    Columns correspond to representative validation trajectories from the
    early, middle, and late stages of training. The three rows show the first,
    middle, and last meta-steps of each trajectory. Each donut chart contains
    three concentric rings corresponding to workflow layers, and each sector
    denotes one candidate agent. Black outlines mark the highest-probability
    agent in each layer.
    }
    \label{fig:adapter_policy_evolution}
\end{figure}

\paragraph{Validation-time adapter behavior.}
To further inspect what the Workflow Adapter learns during training, we
visualize its layer-wise selection probabilities on representative validation
trajectories from the early, middle, and late stages of training.
Figure~\ref{fig:adapter_policy_evolution} shows the adapter probabilities at
the first, middle, and last meta-steps of each trajectory. Each donut chart
contains three concentric rings corresponding to the three workflow layers,
and each sector denotes one candidate agent. The black outline marks the
highest-probability agent in each layer. The visualization provides qualitative
evidence that the adapter develops more structured and state-dependent
selection patterns over training, while still maintaining a soft distribution
over candidate agents.

% ============================================================
\section{Limitations}
\label{app:limitations}

EvoMAS studies execution-time workflow construction over a fixed candidate
agent pool and a fixed maximum workflow depth. This controlled setting enables
fair comparison with automated MAS design baselines, but does not yet cover
open-ended agent creation, dynamic expansion of the agent pool, or adaptive
runtime depth selection. Scaling EvoMAS to larger pools and variable-depth
workflows is an important future direction.

The main training objective uses sparse, verifiable terminal task success.
While this avoids directly optimizing against potentially biased evaluator
judgments, terminal reward can be too sparse in very-hard regimes. Our PRM
analysis shows that evaluator-based process reward can help in such cases, but
richer credit-assignment methods, value baselines, and calibrated process
rewards remain open problems.

Our evaluation covers GAIA, HLE, DeepResearcher, and BrowserComp, but some
high-cost benchmarks are evaluated on limited held-out subsets. Future work
should test larger benchmark splits, longer execution horizons, and more
diverse tool-use environments. In addition, EvoMAS inherits limitations of
LLM-based agent systems, including sensitivity to backbone capability, prompt
templates, tool interfaces, and evaluator reliability.

\end{document}